# An Efficient Target Detection and Recognition Method in Aerial Remote-sensing Images Based on Multiangle Regions-of-Interest

Guangcun Shan*
*School of Instrumentation Science and Optoelectronics Engineering*
*Beihang University*
Beijing 100191, China
*gcshan@buaa.edu.cn

Hongyu Wang
*School of Instrumentation Science and Optoelectronics Engineering*
*Beihang University*
Beijing 100191 China
why0706@buaa.edu.cn

Wei Liang
*School of Electronic, Electrical and Communication Engineering,*
*University of Chinese Academy of Sciences,*
Beijing, China
wliang@mail.ie.ac.cn

Congcong Liu
*Yantai Yundu Haiying UAV Application Technology Co., Ltd.*,
Yantai, China
liucong0610@163.com

Qizi Ma
*School of Instrumentation Science and Optoelectronics Engineering*
*Beihang University*
Beijing, China
1332342914@qq.com

Quan Quan
*School of Automation Science and Electrical Engineering*,
Beihang University
Beijing 100191, China
qq_buaa@buaa.edu.cn

***Abstract*—Recently, deep learning technology have been extensively used in the field of image recognition. However, its main application is the recognition and detection of ordinary pictures and common scenes. It is challenging to effectively and expediently analyze remote-sensing images obtained by the image acquisition systems on unmanned aerial vehicles (UAVs), which includes the identification of the target and calculation of its position. Aerial remote sensing images have different shooting angles and methods compared with ordinary pictures or images, which makes remote-sensing images play an irreplaceable role in some areas. In this study, a new target detection and recognition method in remote-sensing images is proposed based on deep convolution neural network (CNN) for the provision of multilevel information of images in combination with a region proposal network used to generate multiangle regions-of-interest. The proposed method generated results that were much more accurate and precise than those obtained with traditional ways. This demonstrated that the model proposed herein displays tremendous applicability potential in remote-sensing image recognition.**

***Keywords*—*remote sensing image, target detection, target recognition, multiangle regions-of-interest***

## I. Introduction

Given the rapid advancement of remote-sensing technologies, the number of high-resolution remote-sensing images acquired by image acquisition systems on unmanned aerial vehicles (UAVs) or high performance satellites has been increasing, which plays an extremely important role in environmental monitoring, national defense, civil airborne navigation, and in other various fields and applications. Correspondingly, the spectral characteristics of the objects are much more abundant; the spectral differences in the same types of objects are increased, and the spectral differences between the classes are reduced. The appearance of a large number of feature details and the complications of the spectral features in the images led to the reduction of classification accuracy for the spectral–statistical-feature-based traditional methods, such as minimum distance, maximum likelihood, and K-means clustering methods [1]. It is more challenging to effectively and expediently analyze the remote sensing images obtained by the image acquisition systems on UAVs. Compared with the image data acquired by the ground image acquisition system, the image quality acquired by UAV is restricted by the image transmission equipment, and is related to the UAV's speed, attitude, position, and other factors. For example, for a hovering UAV, its jitter will lead to the instability of image video stream, and its flight altitude has a great impact on the image scale. As the flight altitude increases, the image content will increase, the object will become smaller, and the discrimination capacity will decrease. Accordingly, excessively large altitude changes can cause instantaneous switching between the image and video streams. When the UAV flight speed is too high, the image video stream will be switched instantaneously and will be restricted by the image acquisition device, thus causing image blurring. As a consequence, some machine learning algorithms, such as artificial neural networks (ANNs) and support vector machines (SVMs), have been applied for the classification of high-resolution images. In the meantime, interesting features, such as texture and structure, have been added in the classification process [2–5]. Several studies have shown that compared with traditional statistical methods, the multivariate feature image classification based on machine learning algorithms can achieve better results. Here we note that both SVM and ANN are shallow learning algorithms that cannot effectively express complex functions owing to the limited computation unit. Despite the fact that the convolutional neural networks (CNNs) were introduced in the last century, they are still limited by the computing power and the scales of the datasets. However, in recent years, CNNs have achieved outstanding performances due to the availability of powerful graphics processing units (GPUs) that have been used to accelerate matrix operations and the monolithically large databases.

Correspondingly, deep CNNs have achieved outstanding performances for image classification tasks. For example, on the MNIST dataset that is used for digit recognition tasks, the recognition accuracy of CNN now exceeds 99.7%, surpassing the accuracy achieved by humans [5]. In fact, the deep CNNs composed of multilayer nonlinear mapping layers have powerful functional expression capacities, and have high efficiencies in complex classification tasks [6, 7].

Recently, the method of deep learning for object detection and recognition has been widely used in various tasks. There are two main methods in object detection and recognition, which are one-stage method and two-stage method. In this paper, an efficient model based on two-stage method was proposed. In two-stage method, the recognition process is divided into two parts, which are generating and identifying regions of interest. The representative of two-stage method are RCNN[8], spatial pyramid pooling networks(SPP-Nets)[9], fast RCNN[10], and faster RCNN[11]. RCNN was the first two-stage method for object detection and recognition which proposed by Girshick[8]. Compared with the traditional algorithm, this method greatly improves the accuracy of object detection and recognition, and caused significant impact on subsequent studies. Subsequently, based on RCNN, Girshick et al. proposed SPP-Net [9]. The SPP-Net solves the problem of information loss in the process of image's resize. In SPP-Net, a new layer called SPP layer was proposed and it was adopted before the fully connected layer. In this way, it solves the problem of double counting in the convolutional layer's and the SPP-Net can get faster than RCNN in detecting. Then, in 2015, Girshick proposed fast R-CNN [10] that capitalized on the advantages of SPP-Net [9] in which the mapping function of the candidate frame was added so that the network could backpropagate. Another additional improvement of fast R-CNN is a layer called ROI Pooling. This layer can convert feature input of different sizes into feature output of the same size and make it possible for regions of different sizes to adopt the full connection layer with fixed dimensions for classification. The classification layer in fast R-CNN is a softmax layer which different from the SVM layer in RCNN. Subsequently, in 2017, Ren et al. proposed faster R-CNN [11] based on the concept of region proposal networks (RPNs). The advantages of RPNs are a) the regions of interest are extract by sliding window, which have a faster speed than selective search used in fast R-CNN. b) by using a generating methods, the RPNs can get 9 different region proposals and generate thousands of region proposals. The RPNs further improves the speed and accuracy of object detection and recognition.

With the development of two-stage network, object detection and recognition tasks have achieved pretty good results in ordinary images[11,12], but for special images such as remote-sensing images, the recognition effect needs to be improved. The main reason is that compared with ordinary images, remote sensing images have the following difficulties:

*1)* Remote-sensing images are associated with a special shooting perspective. This allows the pretraining model trained with ordinary images to be well suited and applicable for use with remote-sensing images.

*2)* Remote sensing images are acquired by satellites or UAVs at altitudes that exceed kilometers. Thus, large targets will appear very small in size in remote-sensing images. Accordingly, it will be difficult for conventional ANNs to extract the features.

*3)* The number of targets contained in different remote-sensing images is often very uneven. For example, in some scenarios (such as parking lots), a remote sensing image may contain hundreds of target objects (cars). In other scenarios (such as suburban areas), a remote sensing image may have only one object.

*4)* In remote-sensing images, the aspect ratio of some targets that are to be detected is very high (like bridge), which means that ordinary region proposals cannot be effectively detected.

In environmental monitoring and emergency event affairs, aerial remote-sensing images acquired by UAVs can be used to identify the locations of important facilities and important disasters. In civil navigation, remote-sensing images can be used for vehicular road planning. Unlike ordinary photos, remote-sensing images can obtain more information than ordinary images due to their special shooting angles. Therefore, efficient target recognition in remote-sensing images maximizes their advantages. In this study, a new and efficient algorithm is proposed by combining the RPN to generate multiangle regions-of-interest (ROIs) with the ResNet-101 model as the basic CNN model. In the remaining parts, we discuss the details of correlated topic vector (Section II), and then present the experimental results (Section III). Conclusions are drawn in Section IV.

## II. METHOD

### *A. Multilevel CNN*

Herein CNN network is used to extract image features. Given that the first CNN LeNet-5 [13] was proposed by Yann LeCun in 1998, CNNs have undergone many generations of development ever since [14, 15]. The very deep convolutional network (VGG) [15] proposed by Oxford University in 2014, and the Google LeNet [16] that was proposed by Google validated the beneficial effects of small convolution kernels and deepening network depths on target feature extraction. ResNet [12] proposed by Microsoft in 2015 used short-cut connections that deepened the neural network to 100 layers and the gradient can be transferred to the upper layer without loss in the process of back propagation by short-cut to provide comprehensive empirical evidence. This showed that these residual networks are easier to optimize, which can gain accuracy from considerably increased depths. In 2017, Gao Huang and coworkers proposed the DenseNet[17], which further strengthened the transmission of features from the bottom to the top through the structure of Dense Block, and reduced vanishing gradients at the same time.

It is obvious that networks are being developed in a progressively deeper direction following the development of CNNs. The main reason for this is attributed to the fact that as the network layers become deeper, more abstract information can be extracted by networks, which will make the model more conducive to grasping the characteristics of the targets, and thereby have a stronger generalization ability. Nevertheless, with the deepening of the network, the edge information extracted by shallow network will have a great loss, such as the contour of the object, or the position of the object, etc. This is not conducive to the accurate positioning of the target object, especially the small objects. Therefore, the CNN presented in this study adopts the method of multi-layer network output.

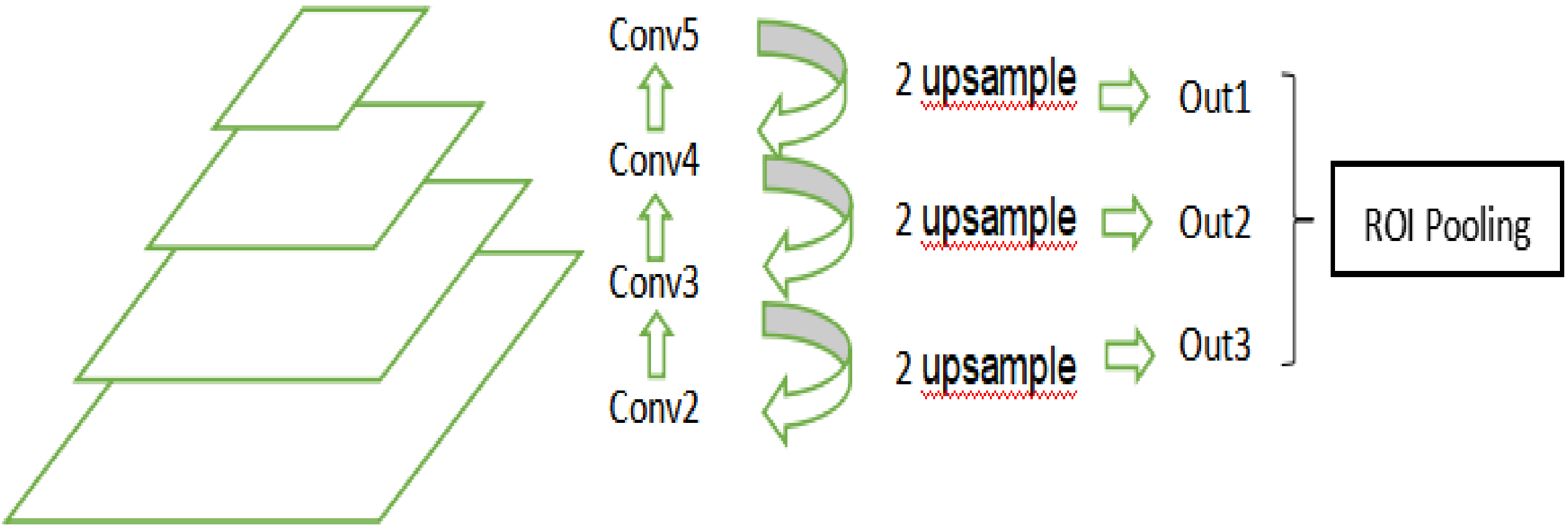

Fig. 1 Model of multi-output convolutional neural networks (CNNs).

In this study, the ResNet-101 model was utilized as the basic model of CNN. In ResNet, the network was divided into five layers, namely, Conv1 to Conv5. The sizes of the feature map outputs of each part were different, that is, the abstraction degrees of the extracted features were different. Conv1 layer extracted the least abstract features, but preserved most of the edge information. Conv5 layer extracted the most abstract features but contained the least amount of edge information. Therefore, to obtain the high-level features of the target and concurrently save adequate edge information, this study used the model structure shown in Fig. 1. In Fig. 1, the outputs of the Conv2, Conv3, Conv4, and Conv5 layers were used for subsequent detection and recognition. The main reason for which the output layer of Conv1 was not used was attributed to the fact that the Conv1 layer feature map was large and too many parameters needed to be calculated. To combine the information between high-level features and low-level features more efficiently, this study refers to the structure proposed by feature pyramid networks (FPNs) [17,18]. The high-level feature map was sampled twice, and the output of the previous network was then added to the feature layer to extract the region proposal in order to make the low-level network have the abstract features of the top-level network and its own edge features with the aim to detect and identify it more accurately.

### B. *Multiangle Region Proposals*

As shown in Fig. 2(a), in order to achieve object detection and recognition in remote-sensing images, the target to be detected is often oriented at a certain angle,. The region proposal given by the traditional target detection and recognition method is rectangular, and will result in the detection effect depicted in Fig. 2(b). It can also be observed that the detection effect is unsatisfactory. Therefore, in this study, the method of multiangle region proposals was utilized to generate the multiangle candidate boxes. With the new method of multiangle region proposals, the candidate box generated here is depicted in Fig. 2(c). As it can be observed, multiangle region proposals are very good at detecting and recognizing targets intended for detection.

Multiangle region recommendation uses the four-point marking method to obtain region proposals. Unlike traditional two-point marker, the four-point marker is constructed by the regression of four coordinates ($x_1$, $y_1$, $x_2$, $y_2$, $x_3$, $y_3$, $x_4$, $y_4$) of an arbitrary quadrilateral. In this way, arbitrary quadrilateral with multiangle can be generated, and the detection and recognition of the detected targets would be treated more appropriately.

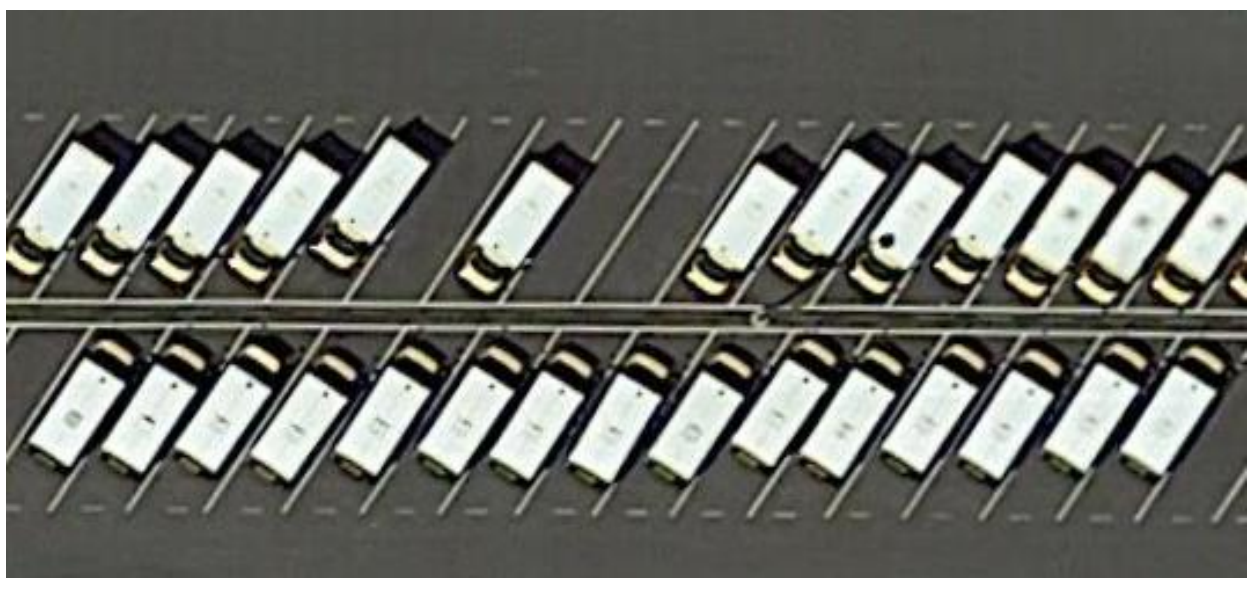

(a)

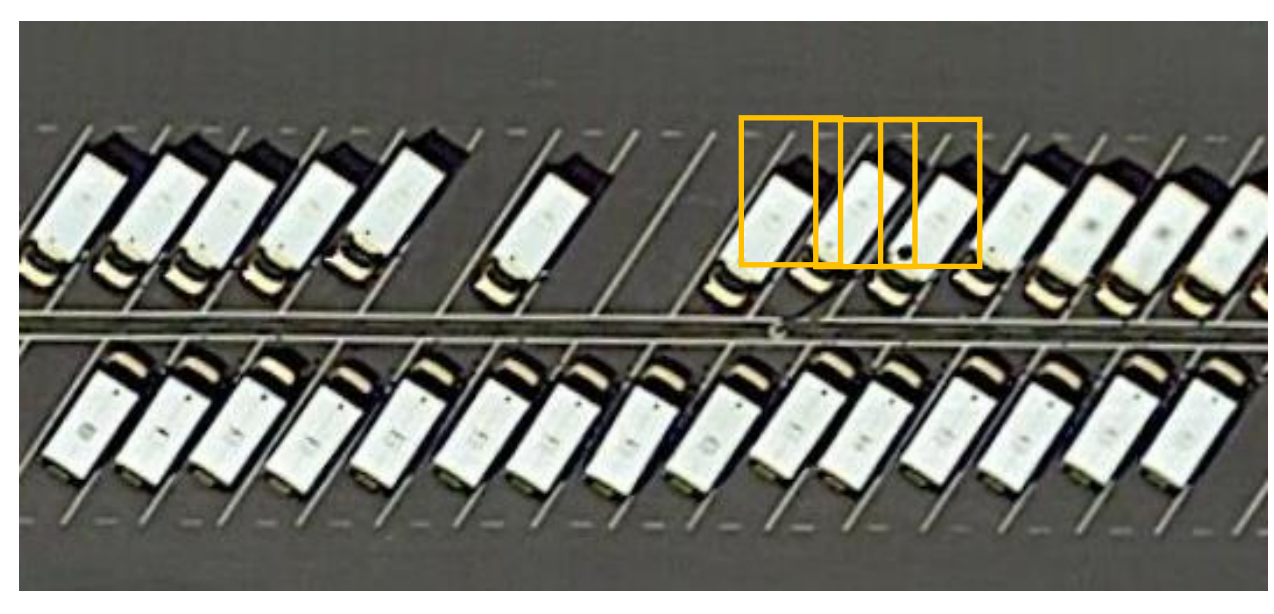

(b)

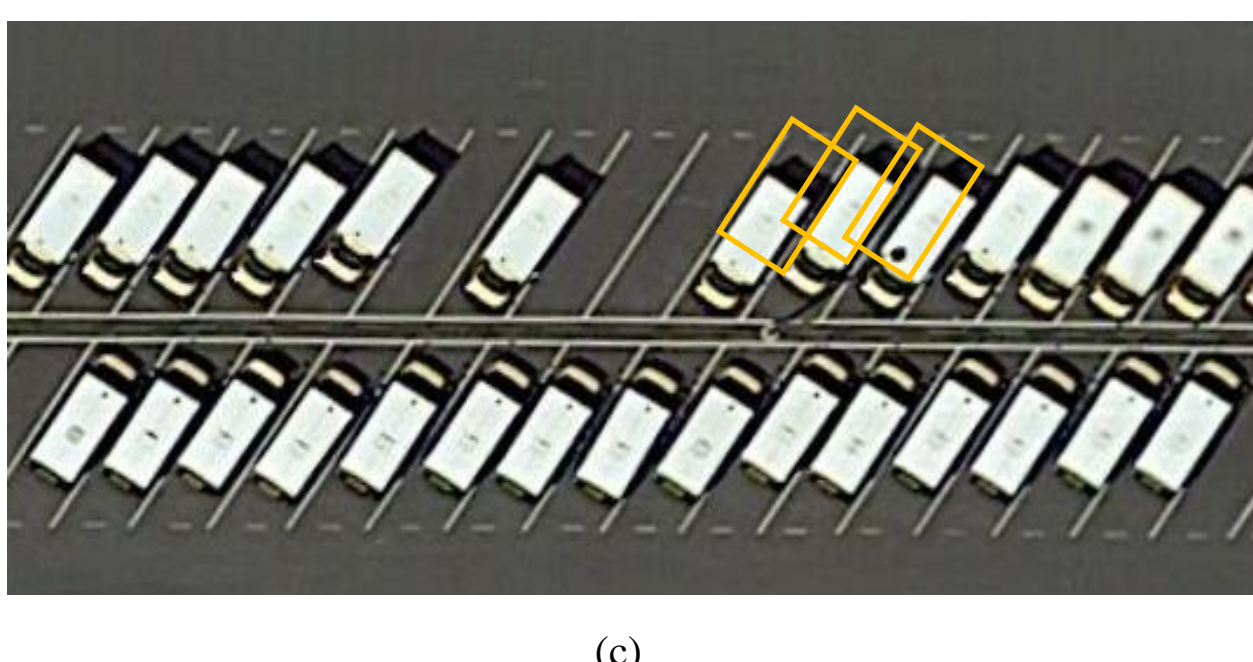

(c)

Fig. 2 Multiangle region proposals. (a) Original picture, (b) rectangular region proposals, and (c) multiangle region proposals.

## III. Experiment and Results

### A. DOTA Database

The target detection in aerial images (DOTA) dataset was used in this work [19]. This dataset was constructed and published in 2018 by Xia and others at Wuhan University, China [19]. The dataset consists of 2806 remote sensing images with a 4000×4000 in-plane resolution and 188282 labeled objects. The objects labeled in the data set include 15 categories: basketball court (BC), baseball diamond (BD), large vehicle (LV), small vehicle (SV), roundabout (RA), storage tank (ST), ground track field (GTF), soccer-ball field (SBF), tennis court (TC), swimming pool (SP), and helicopter (HC). Note that the DOTA dataset is the largest and most complete remote sensing image dataset.

### B. Experiment Results

In the model training experiment, we used Adam as our optimizer with the learning rate set at 0.0001. The base size of the region proposal was 16×16, the anchor scales were 4, 8, 16, 32, 64, and the anchor ratios were 1:1, 1:2, 2:1, 1:8, and 8:1, respectively. The model results are listed in Table I. It is found that the results obtained by using our model are improved greatly compared to the model using faster RCNN only, especially for the small targets and the targets with multiangle characteristics. The results obtained are shown in Fig. 3.

TABLE I. Results for Our Model

| MAP | Plane | BD | Bridge | GTF |
|---|---|---|---|---|
| Faster regions with convolutional neural networks (RCNNs) | **0.802** | **0.696** | 0.096 | 0.559 |
| Our model | 0.755 | 0.404 | 0.372 | 0.463 |
| MAP | **SV** | **LV** | **Ship** | **TC** |
| Faster RCNN | 0.402 | 0.155 | 0.277 | 0.891 |
| Our model | 0.443 | 0.518 | 0.740 | 0.888 |
| MAP | **BC** | **ST** | **SBF** | **RA** |
| Faster RCNN | 0.669 | 0.618 | 0.467 | 0.523 |
| Our model | 0.559 | 0.771 | 0.565 | 0.600 |
| MAP | **Harbor** | **SP** | **HC** | **All** |
| Faster RCNN | 0.178 | 0.449 | 0.334 | 0.474 |
| Our model | 0.582 | 0.487 | 0.336 | 0.565 |

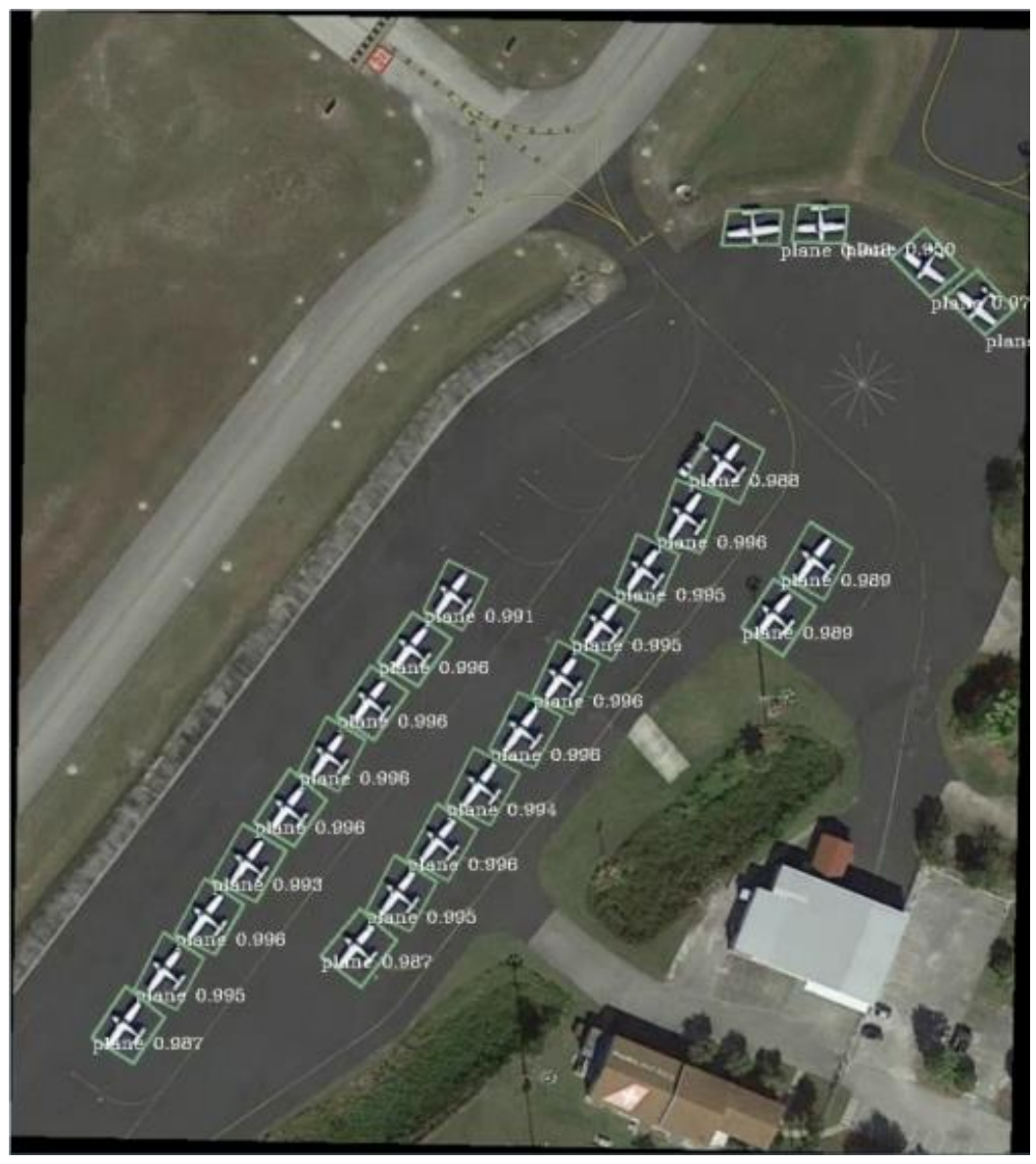

(a)

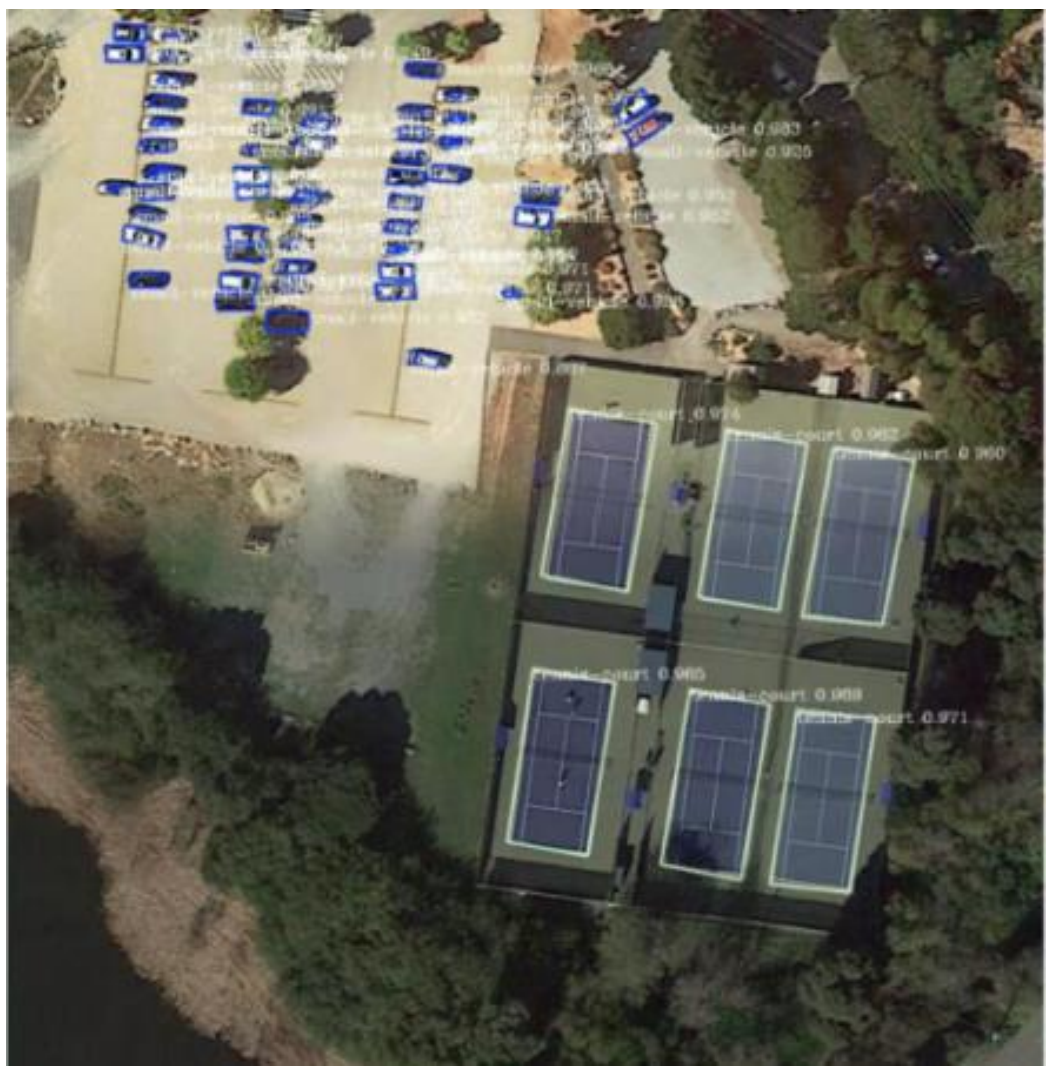

(b)

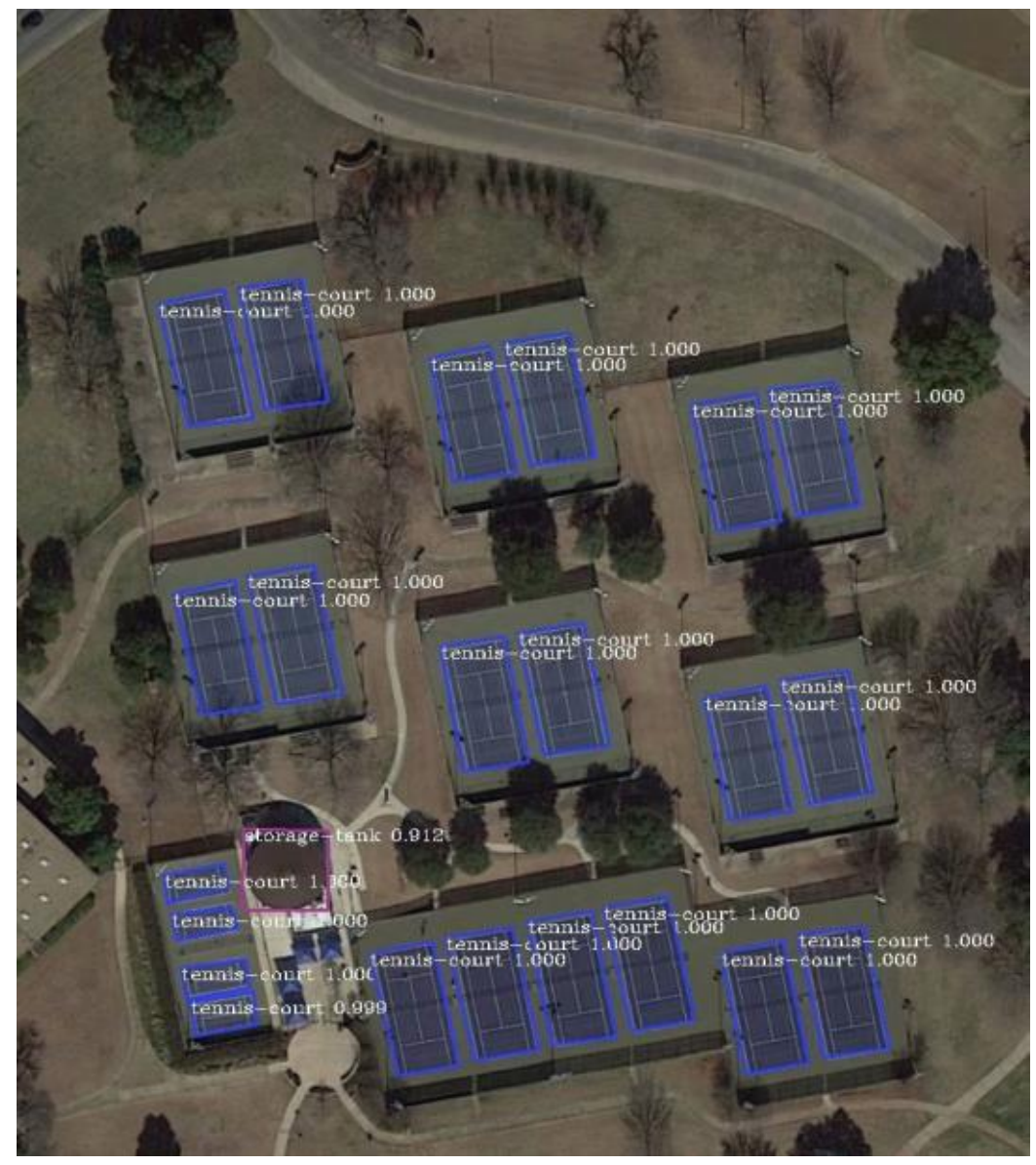

(c)

Fig. 3 Results generated by our proposed model.

## IV. CONCLUSION

This study introduced a new target detection and recognition method in remote-sensing images that yielded more accurate results than those generated by most of the traditional models. The proposed model here used the output of a CNN multilayer to predict the objects of different sizes. At the same time, the four-point marking method was used to generate multiangle region proposals that further improved the accuracy of target detection and recognition. And more deepen networks would be explored in order to achieve better performance.

## ACKNOWLEDGMENT

This work is supported by the National Key Research &Development Program of China (No. 2016YFC1402502).